# CYCLE GENERATIVE ADVERSARIAL NETWORKS ALGORITHM WITH STYLE TRANSFER FOR IMAGE GENERATION

**Anugrah Akbar Praramadhan[1], Guntur Eka Saputra[2]**

Jurusan Sistem Informasi Fakultas Ilmu Komputer dan Teknologi Informasi Universitas Gunadarma[1]

Jurusan Teknik Informatika Fakultas Teknologi Industri Universitas Gunadarma[2]

Jl. Margonda Raya No. 100 Pondok Cina Depok

E-mail : anugrahakbarp@student.gunadarma.ac.id[1], guntur@staff.gunadarma.ac.id[2]

## ABSTRAK

Tantangan terbesar yang dihadapkan kepada seorang *Machine Learning Engineer* yaitu kurangnya data yang dimiliki, khususnya terhadap gambar 2 dimensi. Gambar tersebut diproses untuk dilatih ke dalam suatu model Pembelajaran Mesin agar dapat mengenali pola yang ada pada data tersebut serta memberikan prediksinya. Penelitian ini dimaksudkan untuk membuat solusi dengan menggunakan algoritma *Cycle Generative Adversarial Networks (GANs)* dalam mengatasi masalah kurangnya data tersebut. Kemudian menggunakan *Style Transfer* untuk dapat menghasilkan gambar yang baru berdasarkan style yang diberikan. Berdasarkan hasil uji coba model yang dihasilkan telah dilakukan beberapa improvisasi, sebelumnya nilai loss dari *photo generator: 3.1267, monet style generator: 3.2026, photo discriminator: 0.6325,* dan *monet style discriminator: 0.6931* menjadi *photo generator: 2.3792, monet style generator: 2.7291, photo discriminator: 0.5956,* dan *monet style discriminator: 0.4940.* Penelitian ini diharapkan bahwa penerapan dari solusi ini dapat bermanfaat di bidang Pendidikan, Seni, Teknologi Informasi, Kedokteran, Astronomi, Otomotif, dan bidang penting lainnya.

**Kata Kunci: Pembelajaran Mesin, Kurang Data, Prediksi, Cycle Generative Adversarial Networks, Style Transfer.**

## ABSTRACT

*The biggest challenge faced by a Machine Learning Engineer is the lack of data they have, especially for 2-dimensional images. The image is processed to be trained into a Machine Learning model so that it can recognize patterns in the data and provide predictions. This research is intended to create a solution using the Cycle Generative Adversarial Networks (GANs) algorithm in overcoming the problem of lack of data. Then use Style Transfer to be able to generate a new image based on the given style. Based on the results of testing the resulting model has been carried out several improvements, previously the loss value of the photo generator: 3.1267, monet style generator: 3.2026, photo discriminator: 0.6325, and monet style discriminator: 0.6931 to photo generator: 2.3792, monet style generator: 2.7291, photo discriminator: 0.5956, and monet style discriminator: 0.4940. It is hoped that the research will make the application of this solution useful in the fields of Education, Arts, Information Technology, Medicine, Astronomy, Automotive and other important fields.*

***Keywords: Machine Learning, Lack of Data, Predictions, Cycle Generative Adversarial Networks, Style Transfer.***

**PENDAHULUAN**

     *Machine Learning* atau *Pembelajaran Mesin* adalah kemampuan sebuah mesin yang dapat belajar secara mandiri berdasarkan input yang diberikan tanpa harus berulang kali di program secara eksplisit. Input tersebut merupakan data yang akan dilatih di dalam model *machine learning* dan berguna dalam prediksi pola yang dihasilkan berdasarkan data tersebut. Seringkali beberapa algoritma dari *machine learning* menggunakan data yang tidak sedikit, hal ini membuat suatu tantangan bagi *machine learning engineer* untuk dapat mengolah dan memperoleh hasil yang maksimal dari suatu model machine learning. Beberapa algoritma dari machine learning, seperti *Neural Networks, Convolutional Neural Networks, Recurrent Neural Networks, Support Vector Machine, K-Nearest Neighbors, K-Means, Principal Component Analysis, Deep Q Networks, Deep Deterministic Policy Gradient,* dan *Generative Adversarial Networks*. Banyaknya algoritma yang ada pada *machine learning*, baik itu baru maupun lama memungkinkan dalam pemecahan hampir semua masalah yang ada di era *Revolusi Industri 4.0*.

     *Generative Adversarial Networks (GANs)* adalah salah satu dari algoritma *Unsupervised Learning* yang ada pada *machine learning* sejak tahun 2014, dimana algoritma ini menggunakan 2 *neural networks* atau jaringan syaraf buatan yang terdiri atas *Generator* dan *Diskriminator*. Dengan adanya struktur seperti itu dapat menghasilkan suatu data sintesis baru yang menyerupai bentuk asli pada inputnya. Penggunaan ini sudah cukup luas dalam pembuatan gambar, video, dan suara, sedangkan *Cycle Generative Adversarial Networks* adalah jenis dari algoritma GANs yang memiliki kemampuan dalam *Penerjemahan Gambar* atau *Image Translation*. Perbedaannya pada Cycle GANs data yang dimiliki harus memiliki gambar yang terbagi menjadi *2* bagian atau *domain* untuk dapat dipasangkan, sehingga menghasilkan gambar sintesis yang baru berdasarkan hasil pasangan tersebut. Implementasi dari Cycle GANs ini digunakan untuk *Season Translation, Object Transfiguration, Generating Photos from Paintings,* dan *Style Transfer*.

     Penelitian terkait yang dengan menggunakan algoritma *Cycle Generative Adversarial Networks* yaitu pada tahun 2019 Hao Tang et al dalam penelitiannya mengusulkan *Cycle In Cycle Generative Adversarial Network* (C2GAN) baru untuk tugas pembuatan gambar yang dipandu keypoint. C2GAN yang diusulkan adalah kerangka lintas-modal yang mengeksplorasi eksploitasi bersama dari titik kunci dan data gambar secara interaktif. Hasil eksperimental yang ekstensif pada dua set data yang tersedia untuk umum, yaitu Radboud Faces dan Market-1501, menunjukkan bahwa pendekatan penelitian ini efektif untuk menghasilkan gambar yang lebih realistis foto dibandingkan dengan model mutakhir.

     Pada tahun 2020 Lan Lan et all melakukan penelitian yaitu memperkenalkan asal, prinsip kerja spesifik, dan sejarah pengembangan GAN, berbagai aplikasi GAN dalam pemrosesan citra digital, Cycle-GAN, dan aplikasinya dalam analisis pencitraan medis, serta aplikasi GAN terbaru dalam informatika medis dan bioinformatika. Dari penerapan GAN pada citra medis, dapat dilihat bahwa model berbasis GAN memberikan solusi yang baik untuk kekurangan data dalam analisis citra medis. Ini dapat dianggap sebagai salah satu tambahan penting pada pelabelan manual dari ahli radiologi. Model yang didasarkan pada satu GAN lebih banyak digunakan sebagai metode augmentasi data untuk meningkatkan variasi dan kuantitas gambar dalam modalitas yang sama.

Berdasarkan permasalahan yang ada dan penelitian terkait yang disebabkan oleh *lack of data* atau *kurangnya data*, masalah ini dapat diselesaikan salah satunya menggunakan algoritma *Cycle GANs* dengan harapan data sintesis baru yang dihasilkan oleh algoritma tersebut mampu menandingi performa dari data nyata atau real yang menjadi benchmark awal.

**METODE PENELITIAN**

Sebelum penelitian dimulai, kerangka kerja dari suatu penelitian tersebut harus dibuat terlebih dahulu, agar terstruktur dan sistematis. Kerangka kerja yang dibuat pada gambar 1 dibawah berdasarkan apa yang akan dilakukan pada saat penelitian hingga tahap akhir.

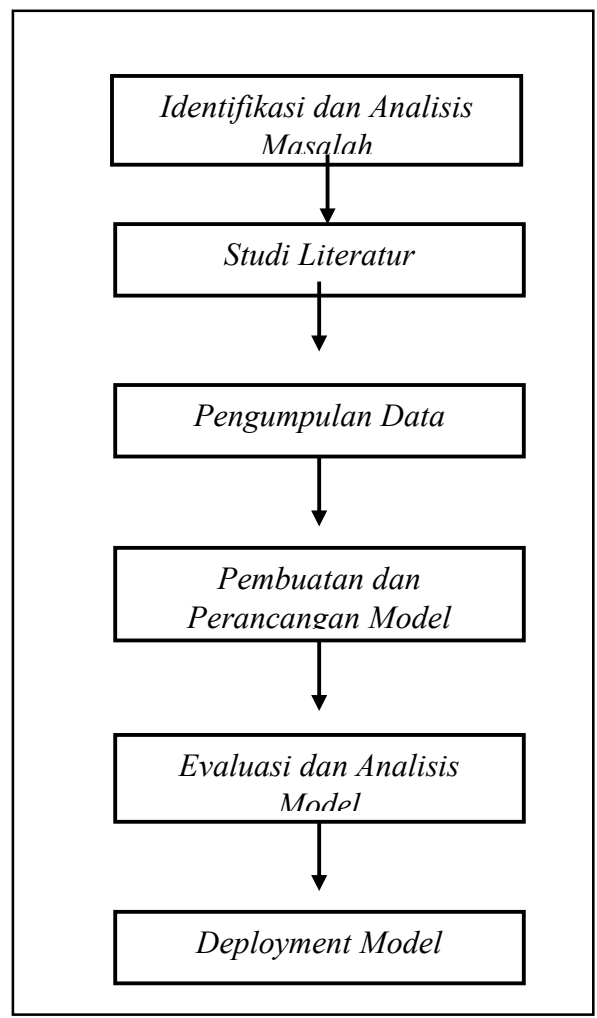

Gambar 1. Kerangka Kerja Metode Penelitian

Berdasarkan kerangka kerja pada gambar 1, tahap pertama yaitu identifikasi dan analisis masalah. Pada tahap ini dilakukan pengamatan terhadap pola perilaku dari Machine Learning Engineer dan beberapa kasus yang dihadapi. Berdasarkan pengamatan bahwa Machine Learning Engineer hampir semua melakukan proses *Deep Learning* dalam penyelesaian kasusnya, dimana menggunakan data yang tidak sedikit untuk pelatihan ke dalam suatu model machine learning.

Tahap kedua, melakukan studi literatur untuk mencari referensi, pengetahuan, fakta, dan beberapa implementasi yang sudah ada terkait penelitian ini, baik itu melalui media cetak seperti buku dan jurnal, maupun melalui media elektronik (non-cetak) seperti e-book, e-paper, dan situs web resmi.

Tahap ketiga, pada tahap ini pengumpulan data dilakukan agar mendapatkan data yang dapat dijadikan bahan untuk diolah lebih lanjut. Untuk data yang digunakan sendiri menggunakan datasets kompetisi yang ada pada platform *Kaggle*.

Tahap keempat, arsitektur yang digunakan pada model yang ingin dibuat yaitu menggunakan arsitektur dari Cycle Generative Adversarial Networks untuk Style Transfer, yaitu kemampuan komputer untuk melakukan transfer style dari satu objek kepada objek tertentu lainnya dalam hal ini berupa foto dua dimensi, kemudian menggabungkan keduanya menjadi satu kesatuan yang berbeda.

Tahap kelima, evaluasi yang dilakukan pada tahap ini bertujuan untuk menganalisis bagaimana performa yang dihasilkan dari model yang telah di buat sebelumnya dengan melakukan tinjau ulang pada model. Jika merasa perlu ditingkatkan terutama dibagian akurasi model, maka kembali ketahap sebelumnya, pengulangan ini terus dilakukan hingga menghasilkan model terbaik berdasarkan data yang ada.

Tahap keenam, model yang telah dievaluasi kemudian ditetapkan menjadi bench mark untuk disebarluaskan ke tahap produksi massal / lanjut.

**PEMBAHASAN**

Cycle Generative Adversarial Networks Architecture, seperti yang sudah dibahas sebelumnya pada penelitian ini menggunakan Cycle GANs sebagai arsitektur yang dipakai. Berikut merupakan bentuk arsitektur dari Cycle GANs:

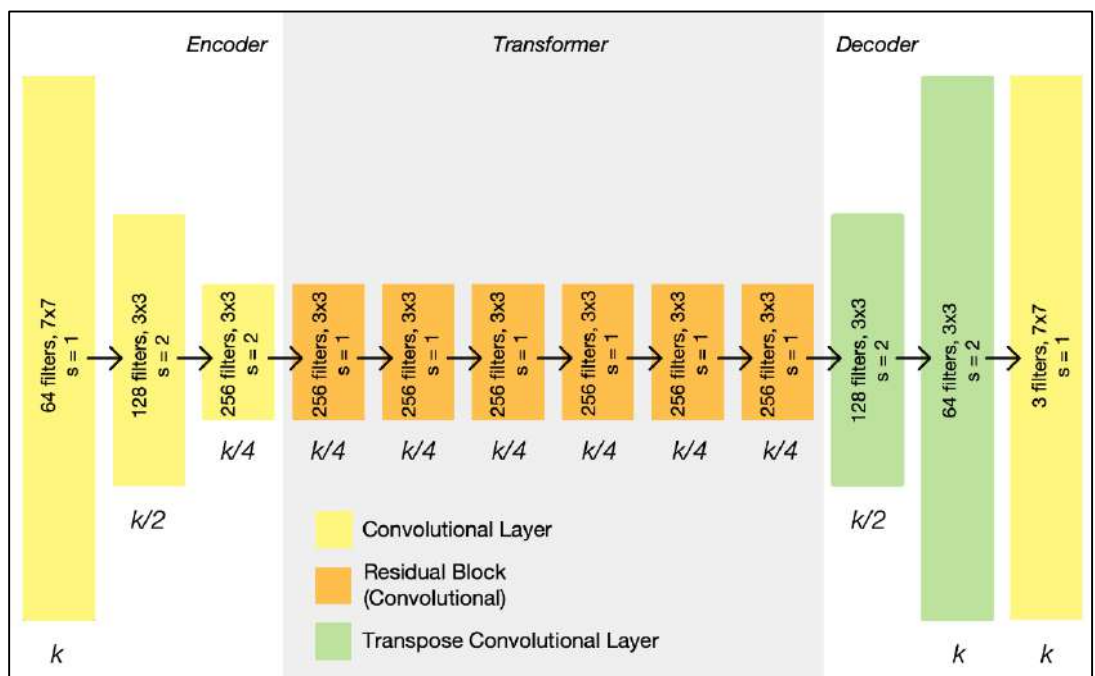

Gambar 2. Cycle GANs Generator Layer Architecture
Sumber : towardsdatascience.com

Gambar 2 merupakan arsitektur dari *layer generator* yang ada pada Cycle GANs. Dapat dilihat bahwa fase *encoder* memiliki penyusutan representasi ukuran yang dihasilkan, konstan pada fase *transformer*, kemudian mengembang pada fase *decoder*. Arsitektur generator dari Cycle GANs diatas merupakan *fully convolutional* layer dimana terdapat *filter* dan *strides* tiap layernya. Pada tiap layernya juga dilakukan normalisasi dan menggunakan fungsi aktivasi *ReLU (Rectified Linear Unit)*. Tahap transformer *(features translation)* merupakan tahap dimana gambar sebagai input tersebut di terjemahkan, sedangkan tahap encoder dan decoder berfungsi sebagai perantara pemrosesan sebelum *(feature extraction)* dan setelah *(decode result)* dilakukan penerjemahan terhadap suatu gambar tersebut.

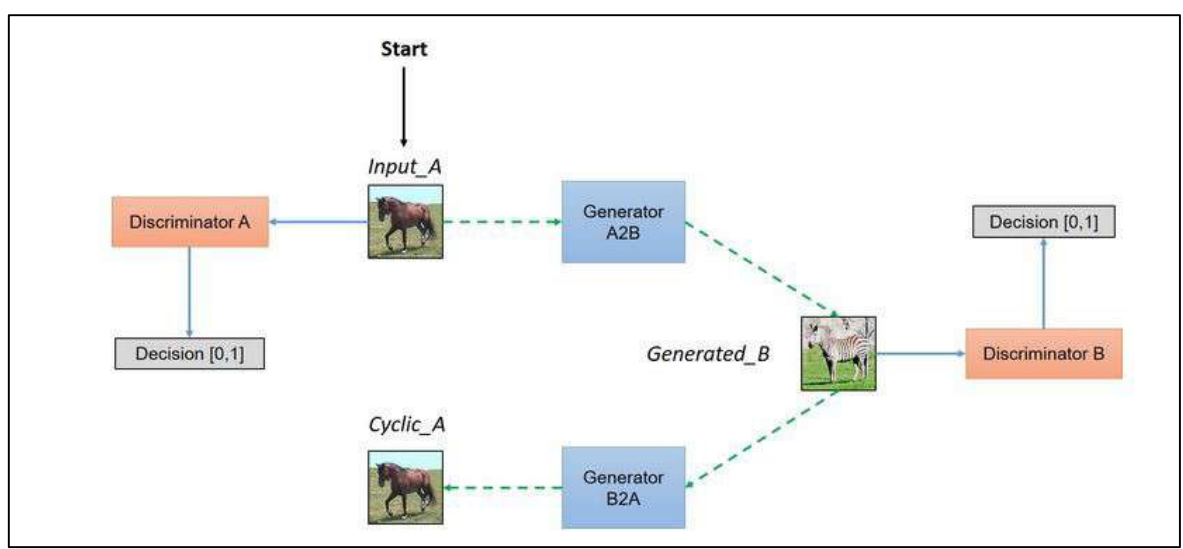

Gambar 3. Cycle GANs WorkFlow Diagram
Sumber : towardsdatascience.com

Pada gambar 3, merupakan gambar untuk diagram alur kerja dari keseluruhan proses Cycle GANs. Dapat dilihat bahwa benar data yang dipakai oleh algoritma Cycle GANs ini harus memiliki 2 buah jenis / domain yang bersifat unpaired / tidak terpasangkan. Kemudian mengacu pada salah satu tujuannya, Cycle GANs dapat memasangkan kedua gambar menjadi 1 gambar baru tanpa harus ada benchmark atau acuan terlebih dahulu. Berdasarkan pada gambar terdiri atas *2 diskriminator*, *2 input gambar*, *2 generator,* dan *1 output gambar*. Diskriminator pertama berfungsi untuk mengecek validasi gambar real yang nantinya akan dibandingkan dengan diskriminator kedua. Pada diskriminator kedua validasi pengecekan diambil berdasarkan gambar yang dihasilkan oleh kedua generator tersebut. Generator pertama untuk gambar asli 1 yang ada pada data, sedangkan generator kedua untuk gambar asli 2 yang ada pada data. Pada generator kedua gambarnya biasanya seringkali digantikan dengan model gambar style, sehingga dinamakan *Style Transfer*. Kemudian kedua generator tersebut menghasilkan satu buah data sintesis berbentuk gambar yang memiliki fitur dari keduanya.

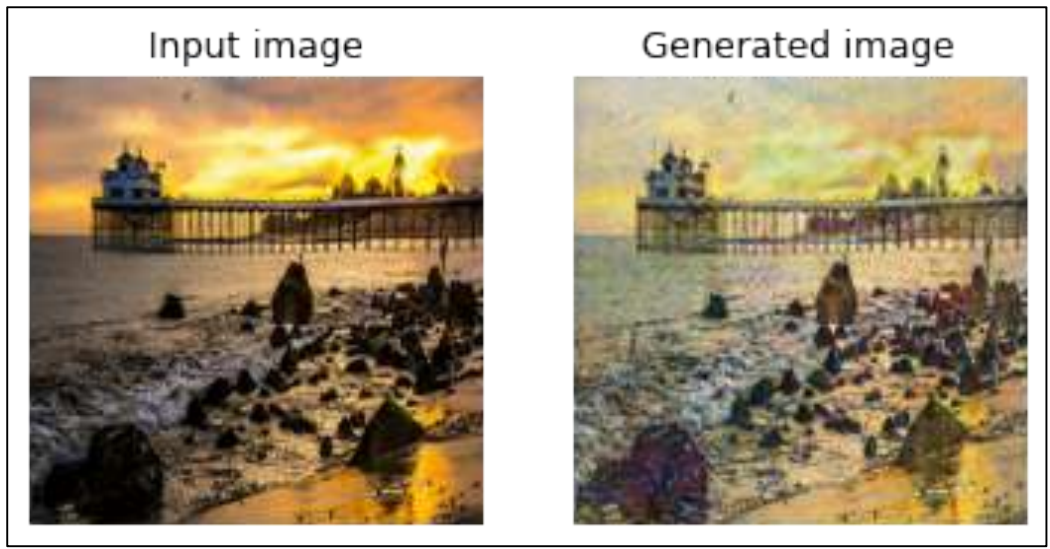

Gambar 4. Image Result of Cycle GANs

Seperti yang diketahui sebelumnya, penelitian ini menggunakan dataset dari platform *Kaggle* sebagai acuan utama. Untuk mengetahui bentuk dari datasets tersebut perlu dilakukan *visualisasi* agar mendapatkan informasi terkait datasets tersebut. Berikut hasil dari visualisasi yang dilakukan :

```
Monet images:
shape: (256, 256, 3)    count: 300
Photo images:
shape: (256, 256, 3)    count: 7038
```

Gambar 5. Total Amounts of Data

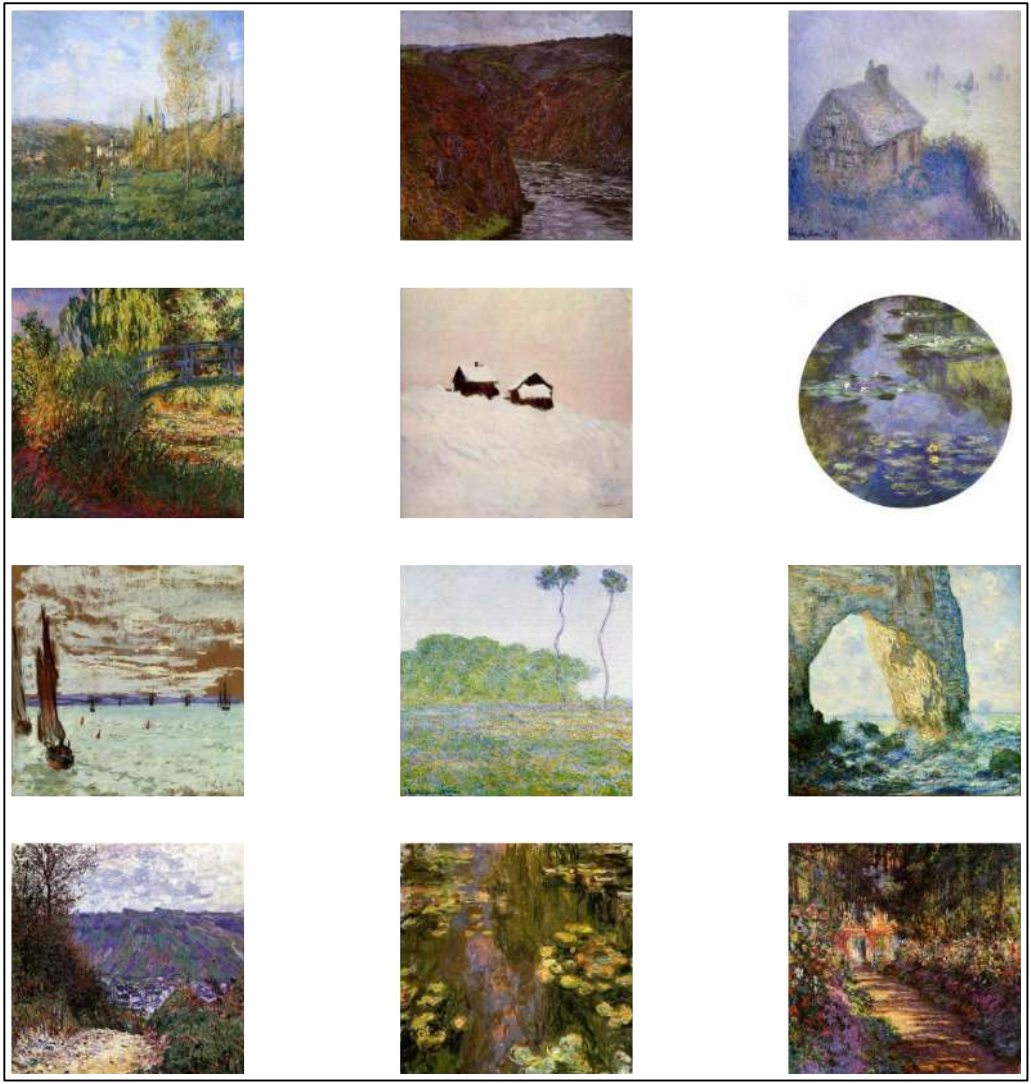

Gambar 6. Monet Style Photos Samples Visualization

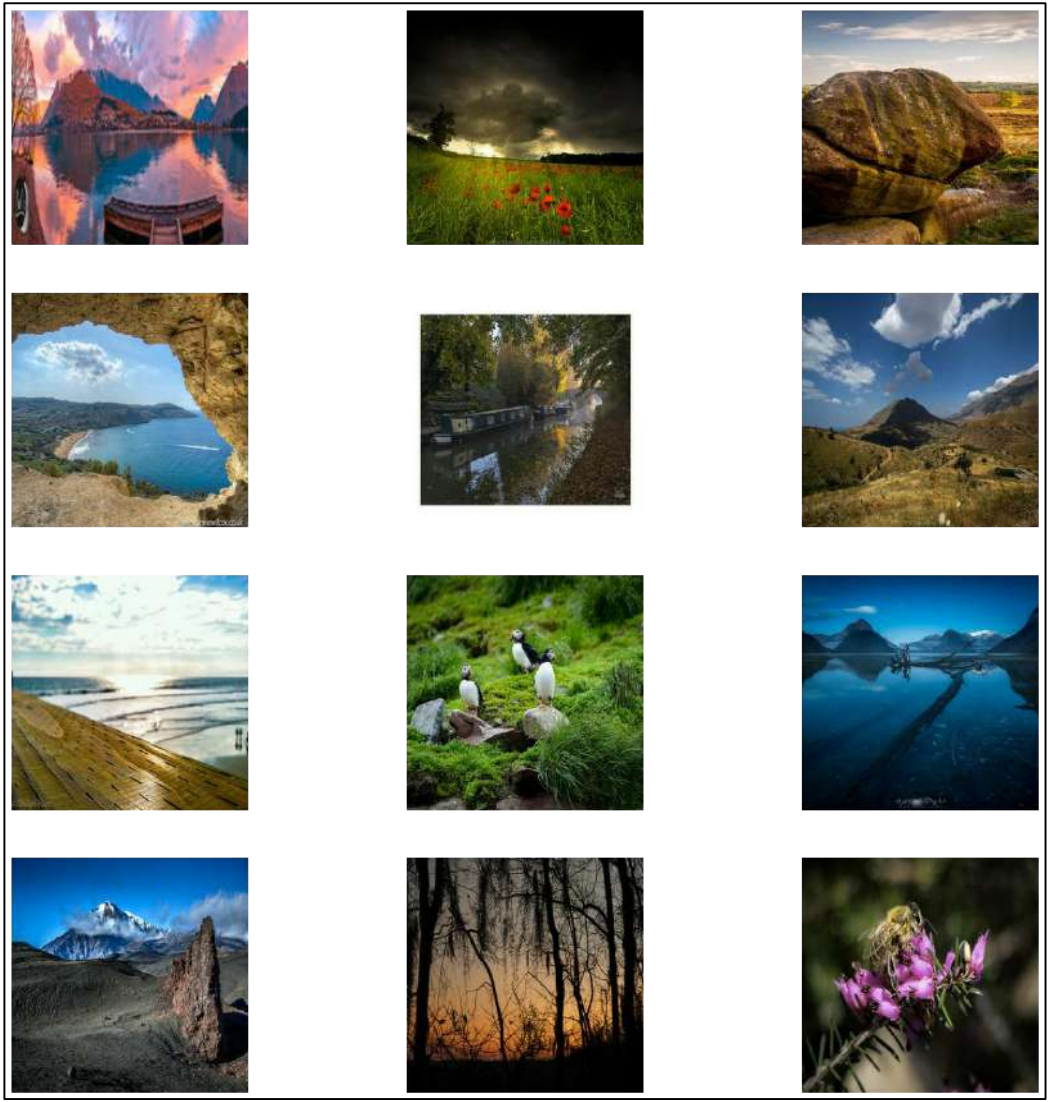

Gambar 7. Base Photos Sample Visualization

Pada gambar 5 diatas, didapatkan informasi berupa ukuran dari kedua kategori gambar tersebut, yaitu *256 × 256 piksel* dalam format RGB. Jika data yang digunakan ukuran tiap gambarnya masih berbeda perlu dilakukan *Image Augmentation* untuk me rescale ukuran tersebut agar sama. Selain itu perbedaan jumlah gambar seperti *300 : 7038* atau *50 : 1173* bukan suatu permasalahan yang besar pada algoritma Cycle GANs karena memang tujuan nya hanya untuk memasangkan kedua gambar tersebut menjadi satu, algoritma tersebut akan otomatis memakai gambar style monet yang sama untuk gambar lainnya jika ada *imbalanced*, sehingga *imbalanced data* tersebut tidak perlu dikhawatirkan. Kemudian pada gambar 6 dan 7 untuk melihat masing-masing gambar berdasarkan dua kategori yang berbeda, untuk dimasukkan ke dalam model dari Cycle GANs.

Setelah gambar-gambar pada tahap sebelumnya siap untuk diproses dan dimasukkan ke dalam model Cycle GANs, langkah selanjutnya yaitu pembuatan model Cycle GANs. Pembuatan model ini mengacu pada struktur arsitekturnya dimulai dari membuat *generator function,* kemudian *discriminator function*, *Cycle GANs function*, dan *loss function* untuk persiapan model sebelum masuk ke tahap training.

```python
def Generator():
    inputs = layers.Input(shape=[256,256,3])

    # bs = batch size
    down_stack = [
        downsample(64, 4, apply_instancenorm=False), # (bs, 128, 128, 64)
        downsample(128, 4), # (bs, 64, 64, 128)
        downsample(256, 4), # (bs, 32, 32, 256)
        downsample(512, 4), # (bs, 16, 16, 512)
        downsample(512, 4), # (bs, 8, 8, 512)
        downsample(512, 4), # (bs, 4, 4, 512)
        downsample(512, 4), # (bs, 2, 2, 512)
        downsample(512, 4), # (bs, 1, 1, 512)
    ]

    up_stack = [
        upsample(512, 4, apply_dropout=True), # (bs, 2, 2, 1024)
        upsample(512, 4, apply_dropout=True), # (bs, 4, 4, 1024)
        upsample(512, 4, apply_dropout=True), # (bs, 8, 8, 1024)
        upsample(512, 4), # (bs, 16, 16, 1024)
        upsample(256, 4), # (bs, 32, 32, 512)
```

Gambar 8. Generator Architecture

*BUILD THE DISCRIMINATOR*

```python
def Discriminator():
    initializer = tf.random_normal_initializer(0., 0.02)
    gamma_init = keras.initializers.RandomNormal(mean=0.0, stddev=0.02)

    inp = layers.Input(shape=[256, 256, 3], name='input_image')

    x = inp

    down1 = downsample(64, 4, False)(x) # (bs, 128, 128, 64)
    down2 = downsample(128, 4)(down1) # (bs, 64, 64, 128)
    down3 = downsample(256, 4)(down2) # (bs, 32, 32, 256)

    zero_pad1 = layers.ZeroPadding2D()(down3) # (bs, 34, 34, 256)
    conv = layers.Conv2D(512, 4, strides=1,
                        kernel_initializer=initializer,
                        use_bias=False)(zero_pad1) # (bs, 31, 31, 512)

    norm1 = tfa.layers.InstanceNormalization(gamma_initializer=gamma_init)
    (conv)

    leaky_relu = layers.LeakyReLU()(norm1)

    zero_pad2 = layers.ZeroPadding2D()(leaky_relu) # (bs, 33, 33, 512)
```

Gambar 9. Discriminator Architecture

*BUILD THE CYCLE GANs MODEL*

```python
class CycleGan(keras.Model):
    def __init__(
        self,
        monet_generator,
        photo_generator,
        monet_discriminator,
        photo_discriminator,
        lambda_cycle=10,
    ):
        super(CycleGan, self).__init__()
        self.m_gen = monet_generator
        self.p_gen = photo_generator
        self.m_disc = monet_discriminator
        self.p_disc = photo_discriminator
        self.lambda_cycle = lambda_cycle

    def compile(
        self,
        m_gen_optimizer,
        p_gen_optimizer,
        m_disc_optimizer,
        p_disc_optimizer,
        gen_loss_fn,
        disc_loss_fn,
```

Gambar 10. Cycle GANs Model

*LOSS FUNCTIONS and COMPILE*

```python
    # Generator loss
    def generator_loss(generated):
        return losses.BinaryCrossentropy(from_
    s.Reduction.NONE)(tf.ones_like(generated), gene
```

Gambar 11. Generator Loss Function

```python
cycle_gan_model.compile(
    m_gen_optimizer = monet_generator_
    p_gen_optimizer = photo_generator_
    m_disc_optimizer = monet_discrimin
    p_disc_optimizer = photo_discrimin
    gen_loss_fn = generator_loss,
    disc_loss_fn = discriminator_loss,
    cycle_loss_fn = calc_cycle_loss,
    identity_loss_fn = identity_loss
)
```

Gambar 12. Model Compile

Model telah dibuat, kemudian akan dilakukan training pada model untuk mendapatkan *loss terkecil* dari masing-masing kategori, yaitu generator dan discriminator berdasarkan gambar foto dan gambar monet style. Berikut merupakan hasil dari model yang sudah di-*training*:

```
Epoch 30/30
1759/1759 [==============================] - 228s 130ms/step - photo_gen
_loss: 2.3792 - photo_disc_loss: 0.5956 - monet_gen_loss: 2.7291 - monet
_disc_loss: 0.4940

<tensorflow.python.keras.callbacks.History at 0x7fc7d426a650>
```

Gambar 13. Train the Model of Cycle GANs

Model yang dihasilkan telah dilakukan beberapa improvisasi, sebelumnya nilai loss dari *photo generator: 3.1267, monet style generator: 3.2026, photo discriminator: 0.6325,* dan *monet style discriminator: 0.6931* menjadi *photo generator: 2.3792, monet style generator: 2.7291, photo discriminator: 0.5956,* dan *monet style discriminator: 0.4940.* Mengacu data tersebut peningkatan performa dari loss berdasarkan angka memang terlihat tidak terlalu signifikan, tetapi jika mengacu pada hasil akhir perbandingan antara model pertama *(simplified)* dan model kedua *(complex)* akan terlihat cukup jelas gambar sintesis yang dihasilkan.

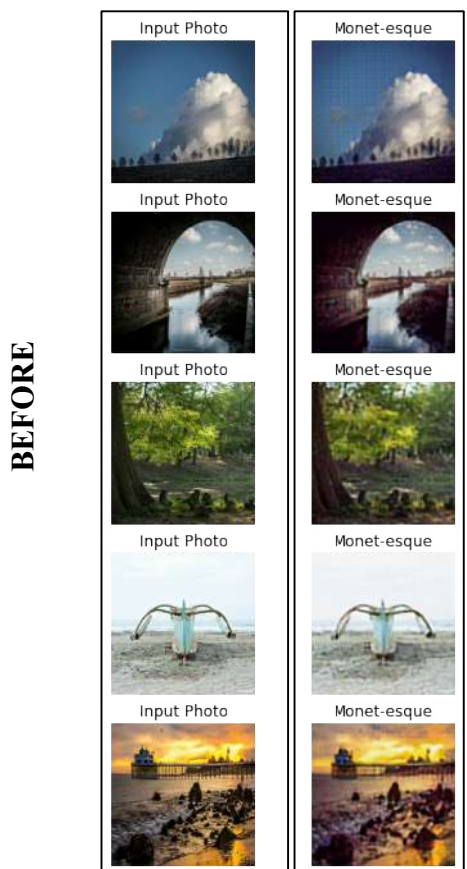 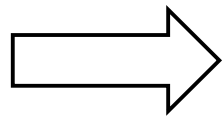 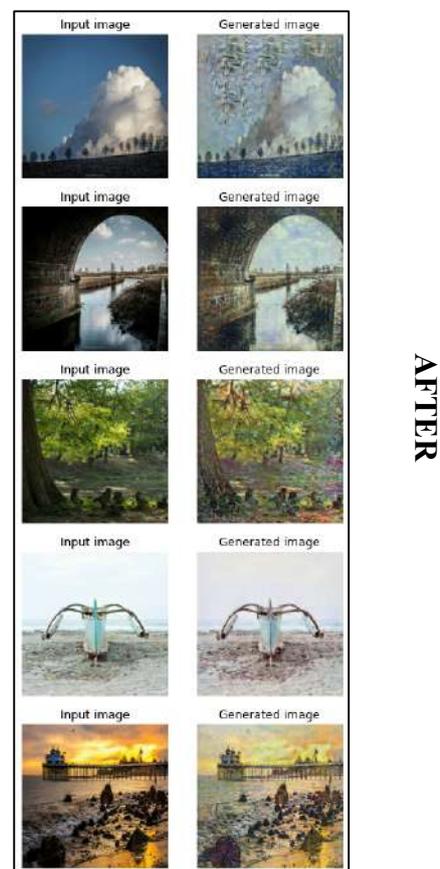

Gambar 14. Result of Simplified Model       Gambar 15. Result of Complex Model

**SIMPULAN DAN SARAN**

Berdasarkan penelitian yang telah dilakukan, algoritma Cycle GANs terbukti memiliki performa yang cukup baik untuk melakukan suatu Style Transfer dan menghasilkan data sintesis baru yang mirip data aslinya, itulah mengapa banyak luas digunakan juga untuk menangani permasalahan kekurangan data. Dalam pembuatan model yang perlu diperhatikan ekstra yaitu saat membuat bagian *generator* dan *loss function* karena itu yang sangat menentukan performa dari akurasi model. Berdasarkan hasil uji coba dalam penelitian ini dihasilkan nilai perbedaan dari nilai sebelumnya yaitu *photo generator: 2.3792, monet style generator: 2.7291, photo discriminator: 0.5956,* dan *monet style discriminator: 0.4940*. Dari hasil tersebut berdasarkan pada hasil akhir perbandingan antara model pertama *(simplified)* dan model kedua *(complex)* terlihat cukup jelas perbedaan gambar sintesis yang dihasilkan dari gambar sebelumnya. Penelitian ini diharapkan dengan adanya kemunculan algoritma-algoritma baru dapat mendukung perkembangan dan pertumbuhan di bidang tersebut agar dapat bersaing dengan negara internasional.

**DAFTAR PUSTAKA**

A Beginner's Guide to Generative Adversarial Networks (GANs),
    URL : https://wiki.pathmind.com/generative-adversarial-network-gan/. Diakses pada: 10 Desember 2020 pukul 21:25.
A Gentle Introduction to Generative Adversarial Networks (GANs),
    URL : https://machinelearningmastery.com/what-are-generative-adversarial-networks-gans/. Diakses pada: 10 Desember 2020 pukul 21:50.
A Gentle Introduction to Cycle Consistent Adversarial Networks,
    URL : https://towardsdatascience.com/a-gentle-introduction-to-cycle-consistent-adversarial-networks-6731c8424a87. Diakses pada: 11 Desember 2020 pukul 17:11.
A Gentle Introduction to CycleGAN for Image Translation,
    URL : https://machinelearningmastery.com/what-is-cyclegan/. Diakses pada: 11 Desember 2020 pukul 19:29.
Chollet, François. 2017. *Deep Learning with Python*. New York: Manning Publications Co.
Complete Repository of CycleGAN by junyanz,
    URL : https://github.com/junyanz/CycleGAN. Diakses pada: 12 November 2020 pukul 19:56.
CycleGAN: Learning to Translate Images (Without Paired Training Data),
    URL : https://towardsdatascience.com/cyclegan-learning-to-translate-images-without-paired-training-data-5b4e93862c8d. Diakses pada: 12 November 2020 pukul 18:46.
Géron, Aurélien. 2019. *Hands-on Machine Learning with Scikit-Learn 2^{nd} Edition, Keras & TensorFlow*. California: O'Reilly Media, Inc.
How to Develop a CycleGAN for Image-to-Image Translation with Keras,
    URL : https://machinelearningmastery.com/cyclegan-tutorial-with-keras/. Diakses pada: 12 Desember 2020 pukul 19:19.
Lan Lan,. You, Lei, et al. 2020. *Generative Adversarial Networks ad Its Applications in Biomedical Informatics*. Front Public Health doi: 10.2289/fpubh.2020.00164.
Monet Style Datasets Images,
    URL : https://www.kaggle.com/c/gan-getting-started/data. Diakses pada: 10 Desember 2020 pukul 20:37.

Müller, Andreas.C., dan Guido Sarah. 2016. *Introduction to Machine Learning with Python*. California: O'Reilly Media, Inc.

Tao, Hang, Xu Dan, et al. 2019. *Cycle In Generative Adversarial Networks for Keypoint-Guided Image Generation*. arXiv:1908.00999 Cornel University.